\pdfoutput=1

\documentclass[11pt]{article}

\usepackage{emnlp2021}

\usepackage{times}
\usepackage{latexsym}

\usepackage[T1]{fontenc}

\usepackage[utf8]{inputenc}

\usepackage{microtype}

\newcommand{\ra}{\(\rightarrow\)} 
\usepackage{tikz-dependency} 

\usepackage[normalem]{ulem} 
\useunder{\uline}{\ul}{} 
\usepackage{graphicx} 
\usepackage[inline]{enumitem} 

\usepackage{csquotes}
\usepackage{multirow}

\title{On the Relation between Syntactic Divergence\\and Zero-Shot Performance}

\newcommand{\affa}{{$^{\dagger}$}}
\newcommand{\affb}{{$^{\ddagger}$}}

\author{
Ofir Arviv\affa\thanks{~~Equal contribution. Dmitry Nikolaev's work was undertaken during his post-doc at Stockholm University.} \quad\quad Dmitry Nikolaev\affb\footnotemark[1] \quad\quad Taelin Karidi\affa  \quad\quad Omri Abend\affa \\
  \affa Hebrew University of Jerusalem \quad\quad
  \affb Institute for Natural Language Processing, University of Stuttgart \\
  \texttt{\{ofir.arviv|taelin.karidi|omri.abend\}@mail.huji.ac.il}\\
  \texttt{dnikolaev@fastmail.com}
}

\begin{document}
\maketitle
\begin{abstract}
We explore the link between the extent to which syntactic relations are preserved in translation and the ease of correctly constructing a parse tree in a zero-shot setting. While previous work suggests such a relation, it tends to focus on the macro level and not on the level of individual edges---a gap we aim to address.
As a test case, we take the transfer of Universal Dependencies (UD) parsing from English to a diverse set of languages and conduct two sets of experiments. In one, we analyze zero-shot performance based on the extent to which English source edges are preserved in translation. In another, we apply three linguistically motivated transformations to UD, creating more cross-lingually stable versions of it, and assess their zero-shot parsability. In order to compare parsing performance across different schemes, we perform extrinsic evaluation on the downstream task of cross-lingual relation extraction (RE) using a subset of a popular English RE benchmark translated to Russian and Korean.\footnote{All resources are available at \url{https://github.com/OfirArviv/translated_tacred} and \url{https://github.com/OfirArviv/improving-ud}}
In both sets of experiments, our results suggest a strong relation between cross-lingual stability and zero-shot parsing performance.
\end{abstract}

\section{Introduction}


Recent progress in cross-lingual transfer methods, such as multi-lingual embeddings \citep{devlin2018bert,mulcaire-etal-2019-polyglot}, enabled significant advances in a wide range of cross-lingual natural language processing tasks. The transferred models, however, are not uniformly effective in addressing languages with different grammatical structures, and little is known about the settings under which cross-lingual transfer is more or less effective. 



A prominent way of facilitating transfer of grammatical knowledge from one language to another is through the use of cross-lingual symbolic representation schemes \citep{chen-etal-2017-improved, Chen2018SyntaxDirectedAF, bugliarello-okazaki-2020-enhancing}. Many advances have been made in this area in recent years, most notably the development and quick adoption of Universal Dependencies \citep[UD;][]{nivre-etal-2016-universal}, a cross-lingually applicable scheme that has become the de facto standard for syntactic annotation. 

\begin{figure}
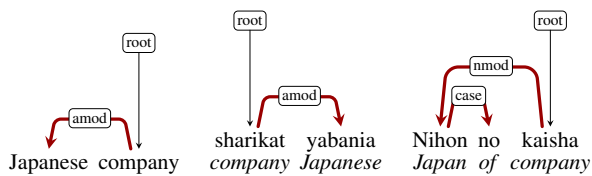

\scalebox{.76}{
\centering
\hspace{-.45cm}
\begin{dependency}
\begin{deptext}
Japanese \& company \\
\end{deptext}
\depedge[edge style={red!60!black,ultra thick}]{2}{1}{amod}
\deproot{2}{root}
\end{dependency}
\begin{dependency}
\begin{deptext}
sharikat \& yabania \\
\textit{company} \& \textit{Japanese} \\
\end{deptext}
\depedge[edge style={red!60!black,ultra thick}]{1}{2}{amod}
\deproot{1}{root}
\end{dependency}
\begin{dependency}
\begin{deptext}
Nihon \& no \& kaisha  \\
\textit{Japan} \& \textit{of} \& \textit{company} \\
\end{deptext}
\depedge[edge style={red!60!black,ultra thick}]{3}{1}{nmod}
\depedge[edge style={red!60!black,ultra thick}]{1}{2}{case}
\deproot{3}{root}
\end{dependency}

}
\begin{small}
\caption{A UD parse of the English phrase \textit{Japanese company} (left) and of its translations to Arabic (middle) and Japanese (right). Word order aside, the UD tree of the Arabic translation is identical to the English one while that of the Japanese translation is different. The {\tt amod} edge in this phrase is stable in translation to Arabic but unstable in translation to Japanese.}
\label{fig:intro_stability_example}
\end{small}
\end{figure}

While these schemes abstract away from many syntactic differences, there is still considerable variability in the strategies employed to express the same basic meanings across languages. In this work, we are mainly interested in the flip side of variability, viz.\ the \textit{stability} of a given scheme---the extent to which its annotations are invariant under translation. See an example in Fig.~\ref{fig:intro_stability_example}.

We present two sets of empirical findings that establish a strong relation between stability and success of zero-shot (ZS) cross-lingual transfer
from a single language in a multiply parallel corpus setting where no annotated data from the target languages were used for training.
Such a setting is a natural starting point for our investigation, as it is both practically useful 
\citep[see, e.g., ][for successful examples of employing ZS learning cross-lingually] {ammar-etal-2016-many,schuster-etal-2019-cross,wang-etal-2019-cross,xu-koehn-2021-zero}
and is based on a homogeneous training set, which minimizes the risk of introducing confounds into the analysis.

The first set of experiments quantifies the effect of stability of edges in UD parse trees on a ZS parser's performance. In order to check if a particular edge was stably transferred from the corresponding source sentence, we use an extended version of the manually aligned subset of the Parallel UD dataset
\citep{zeman-etal-2017-conll,nikolaev-etal-2020-fine}, which provides word-aligned translations of circa 1000 English sentences into six languages with different typological profiles.
We find highly consistent trends across all language pairs, where stable edges receive an average labeled attachment score (LAS; a standard evaluation metric in dependency parsing) that is often twice or more bigger than the one for edges that do not correspond to a source side edge. These findings indicate a strong link between the stability of an edge and its contribution to ZS parsing quality and suggest that ZS parsing performance can be enhanced by improving the stability of the underlying syntactic representation.

Our next set of experiments are the first steps in this direction. 
Concretely, we define three transformations, each targeting an area of syntax that is known to give rise to cross-lingual divergences in UD terms, and thus create three slightly modified versions of UD. We then apply these transformations to the training set in order to check if these versions of UD lead to improved ZS performance. As attachment scores across different schemes are not comparable, we opt for extrinsic evaluation through ZS cross-lingual relation extraction.



Since there are no multilingual RE datasets that include data from non-Western-European languages, we translated a 500-sentence subset of the English TACRED dataset \citep{zhang2017tacred} into Korean and Russian and annotated it with the relations from respective English source sentences. Our results show that modified versions of UD give rise to improved results. This indicates that UD can be made more cross-lingually stable without sacrificing its usefulness for downstream tasks.

The paper is organized as follows.
In \S\ref{sec:edge_stability}, we explore the correlation between ZS performance and stability.
In \S\ref{sec:comparing_schemes}, we present transformations to UD annotations (\S\ref{sec:transformations}) and the methodology for comparing downstream usefulness of vanilla UD and transformed UD (\S\ref{ssec:zs-re}). Our experimental setup is described in \S\ref{sec:model_data}, and the results are presented in \S\ref{sec:results}. Related work is summarised in \S\ref{sec:related_work}. Section \ref{sec:conclusions} concludes the paper.

\section{Edge Stability and ZS Parsability}
\label{sec:edge_stability}

This section evaluates the relation between the extent to which an edge in a translated sentence corresponds to an edge in the original sentence (which we operationalize as several \textit{stability categories}) and the ability of a ZS parser to parse it correctly (its {\it ZS parsability}).
We use the manually aligned subset of the Parallel UD corpus \citep[N20;][]{nikolaev-etal-2020-fine}, which augments the PUD dataset \citep{conll2017shared} with alignments between corresponding content words over five language pairs, the source language being English (En), and the target languages being French (Fr), Russian (Ru), Japanese (Jp), Chinese (Zh), and Korean (Ko). We further use an extension of the corpus to an additional language pair, English-Arabic (En-Ar), in order to increase the diversity of examined languages \citep{rafaeli2021speech}.\footnote{The annotation was carried out by a single annotator, proficient in English and Arabic, using the same guidelines as for the original aligned PUD corpora.}
In all cases, both the UD annotation and the alignment were done manually.

We next train a state-of-the-art ZS parser on En (same as used for the experiment described in \S\ref{ssec:zs-re})  
and examine its performance on PUD over the six target languages. 
We partition the edges in the target-language test parses into categories based on their stability and investigate the performance of the parser on an edge as a function of its stability-category membership.

\subsection{Experimental Setup}\label{lab:edge-stability-setup}

Let $S_e$ be a UD annotated sentence in the source language and $S_l$ its translation in the target language. Let $(w'_1,w'_2)$ be a pair of words in $S_l$ and $(w_1,w_2)$ the corresponding aligned words in $S_e$, if such exist.
We partition the edges in $S_l$ according to the following scheme:


\begin{enumerate}[wide, labelwidth=!, labelindent=0pt]
  \item \textit{Fully Aligned} edges are $e'$=$(w'_1,w'_2)$ in $S_l$ between two content words with label $l$, such that there is an edge  $e=(w_1,w_2)$ is in $S_e$ with label $l$.
  \item \textit{Partially Aligned} edges are $e'=(w'_1,w'_2)$ in $S_l$ between two content words with label $l$, such that there is $e=(w_1,w_2)$ is in $S_e$  with label $l'\neq {l}$.
  \item \textit{Unaligned} edges are $e=(w'_1,w'_2)$ in $S_l$ between two content words where either $w'_1$ or $w'_2$ do not have a single aligned word in $S_e$.
  \item \textit{Flipped} edges are $e'=(w'_1,w'_2)$ in $S_l$ between two content words, such that in $S_e$ there exists an edge $e=(w_2,w_1)$.
  \item \textit{Misaligned} edges are $e'=(w'_1,w'_2)$ in $S_l$ between two content words, where $w'_1$ and $w'_2$ have aligned words in $S_e$ that are not an fully/partially aligned or flipped edges.
  \item \textit{Function Word} edges are $e'=(w'_1,w'_2)$  in $S_l$ where either $w'_1$ or $w'_2$ is a function word.\footnote{Function words were not aligned to other function words in the PUD corpus as such an alignment was deemed unreliable (see N20). Some Function Word edges could be aligned to edges in $S_e$, but this number is low for highly morphosyntactically divergent language pairs.}
\end{enumerate}

We note that each edge belongs to exactly one of the above categories. We report LAS and unlabeled attachment scores (UAS) for each edge category, averaged over 10 runs.
Fully Aligned edges is the most stable category, whereas Misaligned and Flipped edges constitute the least stable one because relying on the edge or path connecting the corresponding words in the original tree would be harmful for performance. Partially Aligned edges represent a special case: this is the second-most stable category in terms of tree structure (UAS) but the least stable one in terms of labels (LAS). 

\begin{table*}[h!]
\small
\centering
\begin{tabular}{|l|cc|cc|cc|cc|cc|cc|}
\hline
\multicolumn{1}{|c|}{} & \multicolumn{2}{c|}{\textbf{Russian}} & \multicolumn{2}{c|}{\textbf{French}} & \multicolumn{2}{c|}{\textbf{Chinese}} & \multicolumn{2}{c|}{\textbf{Japanese}} & \multicolumn{2}{c|}{\textbf{Korean}} & \multicolumn{2}{c|}{\textbf{Arabic}} \\ \cline{2-13} 
\multicolumn{1}{|c|}{Edge Type} & \multicolumn{1}{c|}{UAS} & LAS & \multicolumn{1}{c|}{UAS} & LAS & \multicolumn{1}{c|}{UAS} & LAS & \multicolumn{1}{c|}{UAS} & LAS & \multicolumn{1}{c|}{UAS} & LAS & \multicolumn{1}{l|}{UAS} & \multicolumn{1}{l|}{LAS} \\ \hline
\textbf{Fully Aligned} & 88 & 83 & 88 & 82 & 61 & 48 & 52 & 37 & 54 & 37 & 67 & 50 \\
\textbf{Partially Aligned} & 82 & 41 & 83 & 56 & 56 & 30 & 53 & 24 & 53 & 22 & 63 & 22 \\
\textbf{Unaligned} & 74 & 59 & 78 & 67 & 44 & 28 & 30 & 10 & 41 & 23 & 52 & 32 \\
\textbf{Misaligned} & 54 & 45 & 55 & 49 & 28 & 16 & 22 & 11 & 34 & 13 & 39 & 25 \\
\textbf{Flipped} & 48 & 29 & 54 & 45 & 26 & 13 & 19 & 10 & 37 & 12 & 41 & 23 \\ \hline
\textbf{Func Word} & 68 & 62 & 79 & 72 & 36 & 25 & 20 & 13 & 36 & 20 & 65 & 58 \\ \hline
\end{tabular}
\caption{UD zero-shot performance on the PUD corpora per edge category (averaged over 10 models). Rows correspond to stability categories; columns, to target languages and score types. See \S\ref{sec:edge_stability} for the definitions of stability categories.}
\label{tab:edge_divergence_results}
\end{table*}

\subsection{Results and Discussion}

Our main results are presented in Table \ref{tab:edge_divergence_results}. Standard deviations and the percentages of edge categories can be found in Appendix \ref{sec:edge_category_additional_data}.
We find that the ZS parsability of an edge strongly correlates with its stability. In fact, for UAS, we find that the ZS parsability is almost invariably ordered in the following way:

					

\begin{center}
Fully Aligned $>$ Partially Aligned $>$ Unaligned $>$ Misaligned $>$ Flipped
\end{center}

For LAS, the ordering is similar, except that Partially Aligned is occasionally positioned lower and Misaligned and Flipped swap places in En-Jp.

Moreover, the differences between the scores for the different categories are substantial, with the most stable categories generally obtaining about two times the labeled score of the least stable ones and seeing a $60\%$ increase in unlabeled scores. We can also see a substantial difference between the scores of the Fully Aligned and Unaligned Token edges: around $10$--$20\%$ and $20$--$40\%$ increase in unlabeled and labeled performance, respectively, for structurally similar languages such as En-Ru, and around $30\%+$ (UAS) and $55\%+$ (LAS) increase for more divergent language pairs. This suggests that, despite recent advances in ZS parsing technology, performance is still dependent on the extent to which parallel constructions are prevalent in the source language.

These results lend strong support to our hypothesis that the stability of an edge affects its ZS parsability. They also suggest that when evaluating ZS models, it is informative to distinguish between edges belonging to the most and least stable categories as these categories may pose different challenges that may benefit from the application of different methods. For example, some methods may improve the implicit alignment capabilities of the employed multilingual embeddings while others may improve the parser's ability to abstract away from surface differences and correctly predict unaligned edges.

Function Word edges contain edges of varying stability: some are aligned, while others are not. We thus expect the parser performance on this type of edge to be better than on Flipped and Misaligned edges but lower than on Fully Aligned. Indeed, the results match our expectations, with the exception of LAS in En-Ar, where the LAS on the Function Word edges is higher than that on Fully Aligned edges. In En-Ko and En-Ja, Flipped edges and Unaligned edges, respectively, achieve higher UAS then Function Word edges, but the differences are within one standard deviation.

Standard deviations (over 10 runs) on all edge and score types are small, usually less than 2. The prevalence of each edge type varies depending on the target language's similarity to English, but Partially Aligned and Misaligned edges usually constitute about $4$--$8\%$ each and Flipped Edges about $1.3$--$3\%$, indicating that these difficult cases are present in all languages. For the full data see Appendix~\ref{sec:edge_category_additional_data}.

We note that Partially Aligned edges present considerably lower LAS than Fully Aligned ones. 
Inspecting the Partially Aligned edges that were incorrectly predicted, we see that the parser has a strong bias towards predicting source-side labels. This tendency accounts for $46\%$ and $42\%$ of the errors, respectively, in En-Ru and En-Fr, and for $19\%$, $21\%$, $20\%$ and $31\%$ of the errors, respectively, in En-Zh, En-Ja, En-Ko, and En-Ar. 
These results suggest that defining relations that are less likely to be altered in translation can substantially improve a scheme's transferability, a direction we explore in the following sections.

Last, in order to gain a better insight into the edges constituting each of the stability categories, we analyze the performance of a supervised parser on them (see Appendix~\ref{sec:edge_category_supervised_appendix} for the training setup). We find that, surprisingly, the supervised parsability of an edge strongly correlates with its cross-lingual stability: the parser's performance on different stability categories is generally ordered in the same way as in the ZS setting. This finding suggests that cross-lingual stability has a connection to the ability of the parser to generalize within a language as well, a direction which we defer to future work. 

It must be noted, however, that the performance difference between the categories is less pronounced in the supervised setting than in the ZS setting, which lends support to our hypothesis as to the relation between stability and ZS parsability. For more details, see Appendix~\ref{sec:edge_category_supervised_appendix}. 

To summarize, our analysis shows that ZS parsing performance is better for edges that closely correspond to similar constructions in English. However, while performance increases with stability even for highly similar language pairs (En-Fr and En-Ru), the scores still lag behind their supervised counterparts, which may suggest that the underlying cross-lingual embeddings can be improved.

\section{Comparing Zero-shot Performance across Representation Schemes}
\label{sec:comparing_schemes}

In this section, we aim to manipulate the annotation scheme so as to increase its stability and test whether the modification yields more cross-lingually useful annotations. We achieve this by devising three linguistically motivated, language agnostic transformations (\S\ref{sec:transformations}) and applying them to the downstream task of ZS relation extraction (\S\ref{ssec:zs-re}). While some previous work proposed syntactic preprocessing of the source-side UD trees for the sake of cross-lingual ZS parsing, we are not aware of any previous work that compares the performance on a parsing-dependent downstream task across different schemes. 


\subsection{Transformations}\label{sec:transformations}

We devise three linguistically motivated, universally applicable transformations, based on linguistic typological literature and the findings of N20. The aim of the transformations is to abstract away from syntactic distinctions made by UD that are unstable and of low information value and thus to improve cross-lingual transfer. These transformations bear a conceptual resemblance to the transformations explored in \citep{ponti-etal-2018-isomorphic}. Those, however, are tailored to specific language pairs, while ours are applicable to any language pair. See \S\ref{sec:related_work} for further discussion.

For simplicity, we explore transformations that only alter edge labels and preserve the tree topology 
and defer transformations that alter the tree topology to future work. In terms of the analysis presented in \S\ref{sec:edge_stability}, our transformations aim to increase the congruence between source and target sentences by converting Partially Aligned edges to Fully Aligned ones.

\subsubsection{Normalization of Nominal Modification}

Examining the alignment matrices from N20, we can see that in many languages, \texttt{amod}, \texttt{acl}, \texttt{nmod}, and \texttt{compound} form more or less a complete graph of what they may be aligned with in different languages, which corresponds to observations made by linguists that languages have different patterns of nominal modification \citep{maniez2012corpus,garcia2006mapping}.

We hypothesize therefore that there could be a benefit in representing nominal modification using a simplified, more general scheme. The transformation {\sc Nominal} therefore converts UD \texttt{compound}, \texttt{nmod}, and \texttt{amod} into \texttt{acl}.

\subsubsection{Normalization of Nominal Predicates}\label{ssec:nominal-predicate-normalization}

\begin{figure*}
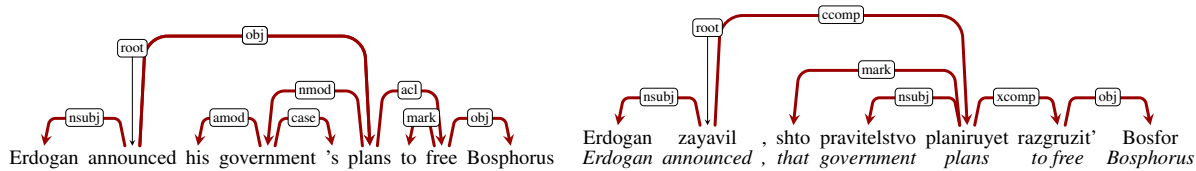

\scalebox{.7}{

\begin{dependency}
\begin{deptext}
Erdogan \& announced \& his \& government \& 's \& plans \& to \& free \& Bosphorus
 \\
\end{deptext}
\depedge[edge style={red!60!black,ultra thick}]{2}{1}{nsubj}
\depedge[edge style={red!60!black,ultra thick}]{2}{6}{obj}
\depedge[edge style={red!60!black,ultra thick}]{4}{3}{amod}
\depedge[edge style={red!60!black,ultra thick}]{4}{5}{case}
\depedge[edge style={red!60!black,ultra thick}]{6}{4}{nmod}
\depedge[edge style={red!60!black,ultra thick}]{6}{8}{acl}
\depedge[edge style={red!60!black,ultra thick}]{2}{6}{obj}
\depedge[edge style={red!60!black,ultra thick}]{8}{7}{mark}
\depedge[edge style={red!60!black,ultra thick}]{8}{9}{obj}
\deproot{2}{root}
\end{dependency}

\begin{dependency}
\begin{deptext}
Erdogan \& zayavil \& , \& shto \& pravitelstvo \& planiruyet \& razgruzit' \& Bosfor \\
\textit{Erdogan} \&  \textit{announced} \& \textit{,} \& \textit{that} \& \textit{government} \& \textit{plans} \& \textit{to free} \& \textit{Bosphorus} \\
\end{deptext}
\depedge[edge style={red!60!black,ultra thick}]{2}{1}{nsubj}
\depedge[edge style={red!60!black,ultra thick}]{2}{6}{ccomp}
\depedge[edge style={red!60!black,ultra thick}]{6}{4}{mark}
\depedge[edge style={red!60!black,ultra thick}]{6}{5}{nsubj}
\depedge[edge style={red!60!black,ultra thick}]{6}{7}{xcomp}
\depedge[edge style={red!60!black,ultra thick}]{7}{8}{obj}
\deproot{2}{root}
\end{dependency}
}
\caption{A pair of sentences from the En-Ru PUD corpus that exemplifies a clausal complement ({\tt ccomp} headed by \textit{planiruyet} in the Russian sentence on the right) that is aligned with a nominal complement of the verb of speech ({\tt obj} headed by \textit{plans} in the English sentence on the left).}
\label{fig:nominal_clausal_example}
\end{figure*}

One of the sources of cross-lingual discrepancies in UD annotations of translated sentences, as detected by N20's data, is the handling of nominal predicates, which UD does not distinguish from other nouns. For example, in \textit{the king's hawk} and \textit{the king's death}, \textit{king} will be labeled as a UD \texttt{nmod} in both cases, although in the first case it is semantically a possessor, while in the second case it is a subject of a change of state. When translated or rephrased, \textit{the king's death} may end up either as a structure that parallels the source or as a verb-headed clause, similar to \textit{the king died}. 
See Figure~\ref{fig:nominal_clausal_example} for a cross-lingual example.



Therefore, UD's lumping of predicative and non-predicative nouns is a potential source of divergence. 
In order to arrive at a more stable handling of predicate nominals, we employ UCCA \citep{abend-rappoport-2013-universal}, a semantic representation that explicitly distinguishes between the two structures.
We use TUPA \citep{hershcovich2018multitask} to parse the training corpus, identify subtrees headed by Processes, and relabel them as subordinate clauses. For more details see Appendix \ref{sec:appendix_transformations}. We use {\sc Predicate} to refer to this transformation.

\subsubsection{Normalization of Obliques}

One of the least clear elements of the UD annotation guidelines is the distinction between \texttt{obl}
(obliques), defined as \enquote{non-core (oblique) arguments or adjuncts}, and \texttt{iobj}, defined as \enquote{any nominal phrase that is a core argument of the verb but is not its subject or (direct) object}. The latter definition
is clarified as referring mostly to arguments of \enquote{ditransitive verbs of exchange}, and the clarification of the former essentially equates is with adjuncts.\footnote{\enquote{This means that it functionally corresponds to an adverbial attaching to a verb, adjective or other adverb.}} This leaves a lot of ambiguous cases (is {\it from his friend} in {\it He got a telephone call from his friend} an
argument of a ditransitive verb of exchange?), and in practice the distinction boils down to whether
there is a case-marking token: PPs are routinely treated as \texttt{obl}, and NPs with non-nominative/accusative
morphological case marking are treated as \texttt{iobj}. 

This leads to a lot of spurious discrepancies between
languages: English corpora usually have only a handful of \texttt{iobj}, while languages with rich case
systems, such as Russian, have them in abundance. Moreover, \texttt{obl} in some languages may often
correspond to \texttt{advmod} in others, which is unsurprising given their overt semantic connection.

In order to bridge these divergences, we propose to retire the \texttt{obl} category altogether, because in most
cases it does not provide any useful information in addition to the fact that there is a nominal-headed subtree with a case-marking token.
We propose to split all \texttt{obl} into \texttt{advmod} and \texttt{iobj} based on their semantics, which we recover by applying the SNACS preposition supersense  parser \citep{Liu2020LexicalSR}\footnote{\url{https://github.com/nelson-liu/lexical-semantic-recognition}} to the input. \texttt{obl} with SNACS tags largely corresponding to typical adverbial semantics\footnote{Concretely: Locus, Time, EndTime, Goal, Source, Purpose,
Duration, Circumstance, ComparisonRef, Manner, Extent.} are converted to \texttt{advmod}, and all others
are converted to \texttt{iobj}.
We use {\sc Oblique} to refer to this transformation.



\subsection{Application to ZS Relation Extraction}\label{ssec:zs-re}

Comparing cross-lingual transferability of different annotation schemes presents a methodological problem: simply comparing the LAS/UAS obtained by a ZS parser may be misleading as the increased performance may come at the expense of losing useful semantic distinctions.

We therefore opt for extrinsic evaluation using the task of ZS cross-lingual relation-extraction on the TACRED dataset \citep{zhang2017tacred}.
Concretely, we take a pattern-based approach to RE, applying the setup that \citet{tiktinsky-etal-2020-pybart} used for monolingual RE to ZS cross-lingual transfer.
RE is a natural choice for this experiment as it is both a core task in NLP, and one that is dependent on syntactic annotations (at least in some state-of-the-art approaches).

Despite the importance of RE, to our knowledge, all existing datasets for this and similar tasks only target Western European languages, which generally resemble English in their grammatical structure. We therefore translate and annotate a little more than 500 examples from the TACRED dataset into Russian and Korean and use these examples as test sets (see Appendix~\ref{sec:tranlated_resources_appendix} for details). We evaluate the performance of the pattern-matching RE models trained on the vanilla and transformed versions of the English TACRED dataset. In order to extract syntactic patterns from the test sets, we use a UD parser trained on the vanilla or one of the transformed versions of the English EWT corpus respectively and apply it to the translated sentences in a ZS setting.

When possible, we also report intrinsic evaluation results, with the caveat mentioned above. We compare the performance of a parser trained on the vanilla English EWT corpus against one that is trained on a transformed version of the corpus. As test sets we use our translated and syntactically annotated subset of the TACRED dataset in Russian and Korean (the vanilla version for the first parser and a transformed version for the second parser).

\section{Experimental Setup}
\label{sec:model_data}

\paragraph{UD Parser.} We use AllenNLP's \citep{gardner-etal-2018-allennlp} implementation of the deep biaffine attention graph-based model of \citet{dozat2017biaffine}. We replace the trainable Glove embeddings with the pre-trained multilingual BERT \citep{devlin2018bert} embeddings\footnote{Specifically, `bert-base-multilingual-cased'.} provided by Hugging Face \cite{wolf-etal-2020-transformers}. We also replace the BiLSTM encoder with self-attention to increase the parser's cross-lingual transfer capabilities \citep{Ahmad2019OnDO}. Finally, we do not use gold (or any) POS tags to represent a more realistic scenario and avoid introducing the potentially confounding factor of POS divergence into our analysis. We use AllenNLP's default settings and hyper-parameters (all hyper-parameter values are given in Appendix~\ref{sec:appendix_models}).

\paragraph{UD Dataset.} We use the UD English-EWT corpus for training our models. We use the standard train-dev-test split and v2.5 of the dataset.

\paragraph{Relation Extraction Model.} We follow the methodology of \citet{tiktinsky-etal-2020-pybart}: for each of the representations, we use the TACRED training set (in English) to acquire extraction patterns. We then apply the pattern set to the test sets in Russian and Korean and report F1 scores.

We explore two settings: a {\sc Standard} one, where we remove the source sentences from which we produced the test set in Korean/Russian from the training set, and a simpler {\sc Parallel} one where these examples are included in the training set. {\sc Parallel} emphasizes the ability of the UD parser to produce similar parses in English and in the target language, while {\sc Standard} emphasizes the parser's ability to generalize from other inputs.

Pattern extraction is performed as follows. Given a labeled sentence equipped with a relation name, participant spans, participant types, and a list of trigger words \citep{yu-etal-2015-read} collected for different relations (see Appendix~\ref{sec:tranlated_resources_appendix} for more details), we find the shortest dependency path between the tokens that connects the entity spans via the trigger words, if such exist.\footnote{In parser outputs, the entity span sometimes does not constitute a sub-tree. Consequently, there may be multiple paths between entity spans and trigger words, of which we select the shortest one.}
More precisely, if a trigger word is found in the sentence, we first compute the shortest paths between the tokens of the first participant and the trigger word and between the trigger word and the tokens of the second participant. We then form an extraction pattern from the path.\footnote{For example: \texttt{PERSON $<$ nsubj "per\_residence" $>$ obj $>$ compound CITY}, where {\tt "per\_residence"} is a type of trigger word.} If no trigger word is found, we compute the shortest path between the two entities and form a pattern in a similar way.\footnote{For example: \texttt{PERSON $<$ nsubj $>$ obj $>$ compound CITY.}} Each pattern is stored in a pattern dictionary together with the number of times it was seen in the training set with a specific relation.

For decoding, we extract the pattern(s) from the input sentence and look up the relations associated with each one in the training pattern dictionary, if such exist. A~majority-vote algorithm is then used for prediction.\footnote{The pattern extraction and evaluation code is partially based on the \href{https://allenai.github.io/pybart}{pyBART} package.}

\section{Results and Discussion}
\label{sec:results}

\paragraph{Extrinsic Evaluation.}

The results are presented in Table \ref{tab:re_results_main}. In the {\sc Parallel} setting, all three transformations display noticeable improvements for both Korean and Russian, increasing both recall and precision. The increase in precision suggests that our transformations not only normalize patterns but also make them more effective.
In the {\sc Standard} setting, all three transformations show improvements for Korean. For Russian, however, only the {\sc Nominal} transformation, which lumps several categories into one, is beneficial.
 
 \begin{table*}[h!]
\small
\centering
\begin{tabular}{|l|ccc|ccc|ccc|ccc|}
\hline
 & \multicolumn{6}{c|}{{\ul \textbf{Standard}}} & \multicolumn{6}{c|}{{\ul \textbf{Parallel}}} \\ \cline{2-13} 
 & \multicolumn{3}{c|}{\textbf{Korean}} & \multicolumn{3}{c|}{\textbf{Russian}} & \multicolumn{3}{c|}{\textbf{Korean}} & \multicolumn{3}{c|}{\textbf{Russian}} \\ \cline{2-13} 
\multicolumn{1}{|c|}{Trans.} & P & R & F1 & P & R & F1 & \multicolumn{1}{l}{P} & \multicolumn{1}{l}{R} & \multicolumn{1}{l|}{F1} & \multicolumn{1}{l}{P} & \multicolumn{1}{l}{R} & \multicolumn{1}{l|}{F1} \\ \hline
\sc{Baseline} & 73.8 & 12.1 & 20.8 & 90.3 & 18.8 & 31.1 & 71.7 & 12.7 & 21.6 & 90.3 & 18.9 & 31.3 \\
\sc{Nominal} & 77.5 & 15.7 & 26.1 & 90.4 & 20.4 & 33.3 & 75.4 & 16.6 & 27.2 & 90.4 & 21.6 & 34.8 \\
 & \textbf{{\ul +3.7}} & \textbf{{\ul +3.6}} & \textbf{{\ul +5.3}} & {\ul +0.1} & \textbf{{\ul +1.6}} & \textbf{{\ul +2.2}} & \textbf{{\ul +3.7}} & \textbf{{\ul +3.9}} & \textbf{{\ul +5.6}} & {\ul +0.1} & \textbf{{\ul +2.7}} & \textbf{{\ul +3.5}} \\
\sc{Predicate} & 75.3 & 12.9 & 22.1 & 89.6 & 18.2 & 30.2 & 75.5 & 13.8 & 23.3 & 90.6 & 21.1 & 34.2 \\
 & {\ul +1.5} & \textbf{{\ul +0.8}} & \textbf{{\ul +1.3}} & {\ul -0.7} & {\ul -0.6} & {\ul -0.9} & \textbf{{\ul +3.8}} & {\ul +1.1} & {\ul +1.7} & {\ul +0.3} & \textbf{{\ul +2.2}} & \textbf{{\ul +2.9}} \\
\sc{Oblique} & 77.8 & 13.1 & 22.4 & 90 & 17.4 & 29.2 & 78.6 & 13.9 & 23.6 & 90.6 & 20.6 & 33.6 \\
 & \textbf{{\ul +4}} & {\ul +1} & \textbf{{\ul +1.6}} & {\ul -0.3} & {\ul -1.4} & {\ul -1.9} & \textbf{{\ul +6.9}} & \textbf{{\ul +1.2}} & \textbf{{\ul +2}} & {\ul +0.3} & \textbf{{\ul +1.7}} & \textbf{{\ul +2.3}} \\ \hline
\end{tabular}
\caption{Experimental results for extrinsic evaluation of the transformed versions of UD on the pattern matching RE task across the {\sc Standard} and {\sc Parallel} settings (see \S\ref{sec:model_data}). Columns correspond to the evaluations settings, target languages, and score types; rows correspond to UD variants. The difference between the scores on the transformed UD corpora and the baseline are shown underlined below the respective scores. Differences with bootstrap $p$-values $< 0.05$ are in boldface. The $p$-values for recall and precision differences for Korean are below $0.08$.}
\label{tab:re_results_main}
\end{table*}

The fact that all three transformations display significant gains in the {\sc Parallel} setting suggests that the UD parser indeed produced similar parses in English and in Russian, as intended. However, the mixed results in the {\sc Standard} setting suggest a difficulty in generalizing across different sentences.

We note that the use of external tools in {\sc Predicate} and {\sc Oblique} adds a considerable amount of noise to the annotation. For example, TUPA, a parser which is used in the second transformation, obtains an F1 score of only around 70 for the relevant labels.
We further note that obtaining improvement for Russian is, on the face of it, more difficult than doing so for Korean as Russian is much more similar to English. Indeed, our analysis in section \S\ref{sec:edge_stability} shows that, in terms of UD, Russian is as close to English as French and the annotation for En-Ru is already very stable.
 
  
Therefore, it seems that in the {\sc Parallel} setting the signal for RE is strong enough to combat the added noise, while in the {\sc Standard} setting, which requires generalization across different sentences and thus uses a weaker signal, results are mixed.
 
 
To validate our hypothesis, we explore an additional setting that can potentially improve the ``signal to noise ratio'': we use both the vanilla and the transformed UD parses, for both the RE train and test sets, so that during training and decoding the RE model can use the patterns that appear in either parses. We denote this setting as the {\sc Ensemble} setting.

The results, reported in Table \ref{tab:re_results_denoise}, are an average of all combinations of outputs of (i)~one out of ten vanilla-trained parsers and (ii)~another vanilla-trained parser (for the baseline) or one out of ten parsers trained on one of the three transformed training sets. 
We thus mitigate the effect of noisy patterns by allowing the model to take recourse to the vanilla patterns in cases where those are more useful.


 \begin{table*}[h!]
\small
\centering
\begin{tabular}{|l|ccc|ccc|ccc|ccc|}
\hline
 & \multicolumn{6}{c|}{{\ul \textbf{Standard -- Ensemble}}} & \multicolumn{6}{c|}{{\ul \textbf{Parallel  -- Ensemble}}} \\ \cline{2-13} 
 & \multicolumn{3}{c|}{\textbf{Korean}} & \multicolumn{3}{c|}{\textbf{Russian}} & \multicolumn{3}{c|}{\textbf{Korean}} & \multicolumn{3}{c|}{\textbf{Russian}} \\ \cline{2-13} 
\multicolumn{1}{|c|}{Trans.} & P & R & F1 & P & R & F1 & \multicolumn{1}{l}{P} & \multicolumn{1}{l}{R} & \multicolumn{1}{l|}{F1} & \multicolumn{1}{l}{P} & \multicolumn{1}{l}{R} & \multicolumn{1}{l|}{F1} \\ \hline
\sc{Baseline} & 71.5 & 21.6 & 33 & 90.1 & 24 & 37.9 & 71.5 & 22.7 & 34.3 & 89 & 24.2 & 38.1 \\
\sc{Nominal} & 75.6 & 29 & 41.8 & 90.3 & 27 & 41.6 & 74.7 & 30.6 & 43.2 & 90.5 & 28.8 & 43.7 \\
 & {\ul +4.1} & {\ul +7.4} & {\ul +8.8} & {\ul +0.2} & {\ul +3} & {\ul +3.7} & {\ul +3.2} & {\ul +7.9} & {\ul +8.9} & {\ul +1.5} & {\ul +4.6} & {\ul +5.6} \\
\sc{Predicate}  & 71.5 & 24.7 & 36.6 & 88.8 & 25.1 & 39.1 & 72.4 & 29.6 & 41.9 & 89.8 & 31.1 & 46.1 \\
 & {\ul 0} & {\ul +3.1} & {\ul +3.6} & {\ul -1.3} & {\ul +1.1} & {\ul +1.2} & {\ul +0.9} & {\ul +6.9} & {\ul +7.6} & {\ul +0.8} & {\ul +6.9} & {\ul +8} \\
\sc{Oblique} & 75.9 & 25.2 & 37.6 & 90.7 & 24.4 & 38.4 & 74.2 & 29.3 & 41.8 & 89.9 & 30.7 & 45.8 \\
  & {\ul +4.4} & {\ul +3.6} & {\ul +4.6} & {\ul +0.6} & {\ul +0.4} & {\ul +0.5} & {\ul +2.7} & {\ul +6.6} & {\ul +7.5} & {\ul +0.9} & {\ul +6.5} & {\ul +7.7} \\ \hline
\end{tabular}
\caption{Experimental results for extrinsic evaluation of transformed UD annotations on the pattern matching relation extracting task, compared against the standard UD in the {\sc Ensemble} setting. Columns and rows are as in Table \ref{tab:re_results_main}. All positive differences are with bootstrap $p$-values $< 0.05$ and the vast majority have $p$ values $< 0.001$.}
\label{tab:re_results_denoise}
\end{table*}

We find significant gains in performance for both Korean and Russian with all three transformations. This lends support to our hypothesis that the mixed results obtained in {\sc Standard} for Russian are due to the noise in the implementation of the transformations and thus should not be interpreted as evidence against our hypothesis as to the relation between stability and ZS parsability.

We use the paired bootstrap test to compute the $p$-values for the the difference between the baseline scores and the different transformations scores. For the {\sc Ensemble} setting, we find that all positive differences are significant ($<0.05$) and the vast majority are highly significant ($<0.001$). For the non-{\sc Ensemble} setting, we find that most are significant ($<0.05$).\footnote{Exact $p$-values can be found in Appendix~\ref{sec:appendix_p_val}.}

\paragraph{Intrinsic Evaluation.}
We report intrinsic evaluation results only for {\sc Nominal}, because the other transformations require parsers for Russian and Korean that are not available to us.

The results, averaged over 10 models, are the following: the vanilla UD parser achieves LAS and UAS of $0.568$ and $0.674$ for Russian and $0.129$ and $0.24$ for Korean, respectively. The transformed UD parser achieves the scores of $0.589$ LAS and $0.676$ UAS for Russian and $0.138$ and $0.24$ for Korean, respectively. That is, the LAS improvement is of $0.021$ and $0.09$ LAS points for Russian and Korean, respectively, and is statistically significant.\footnote{We use the paired bootstrap test to compute the $p$-values for the the difference between the baseline and transformed LAS, all of which are $<0.001$.}
For UAS, the differences are insignificant, which fits in with our discussion in \S\ref{sec:edge_stability}, suggesting that the difference in UAS between Partially Aligned and Fully Aligned edges is small.


\section{Related Work}\label{sec:related_work}

The relation between stability and cross-lingual transfer has been the subject of previous work. However, the great majority of it has focused on word order \cite{wang-eisner-2018-synthetic, rasooli-collins-2019-low, liu-etal-2020-cross-lingual-dependency} and morphological features \cite{wang-eisner-2018-surface, meng-etal-2019-target} and did not address features that entail stronger structural misalignment. \citet{ponti-etal-2018-isomorphic} and \citet{nikolaev-etal-2020-fine} have shown that cross-lingual divergences in UD annotations of constructions with identical semantics have an effect on cross-lingual transfer but did not precisely quantify this effect. In this work, we proposed a stability-category classification of UD edges and then investigated to what extent the performance of a ZS parser on a given edge is affected by its stability in translation, providing insight into the ability of syntax-based ZS models to generalize over different types of divergences.

\citet{ponti-etal-2018-isomorphic} also demonstrated the benefits of modifying UD parse trees in order to improve their utility for cross-lingual applications. These modifications, however, took UD for granted and only targeted specific types of subtrees to make them look more like the corresponding subtrees in the target language (e.g., by converting an English possessive construction \textit{I have X} into a more Arabic-looking locative-possessive construction \textit{is X at me}). This approach, therefore, is highly language-pair specific as the transformations defined on the differences between surface syntax of English and Arabic will not be useful when presented with another target language. Our work, on the other hand, unconditionally altered the scheme itself, and models using it can be profitably transferred to any target language.

Others \citep{stanovsky-etal-2014-intermediary,Schuster2016EnhancedEU, reddy2017universal, nivre-etal-2018-enhancing, tiktinsky-etal-2020-pybart} also proposed transformations aimed at emphasizing useful connections between tokens but not in a cross-lingual context.

\section{Conclusion}
\label{sec:conclusions}

Our work establishes a strong relationship between stability and cross-lingual transfer, even at the level of individual edges. We find that the stability of an edge is a strong indicator for the ability of a ZS parser to predict it, suggesting that despite recent advances in ZS parsing and cross-lingual embeddings, these models still face difficulties in generalizing over the grammars and syntactic-usage patterns. Furthermore, we show that it is possible to improve the annotation stability of representations using linguistically motivated and universally applicable transformations, which lead to better cross-lingual transferability.

Our results suggest several directions for future work in terms of designing more transferable annotation schemes and improving evaluation practices for ZS parsing. They also suggests a path towards a theory of the relation between the linguistic properties of a construction and the ability to effectively process it using cross-lingual-transfer tools.

\section*{Acknowledgments}

This work was supported by the Israel Science Foundation (grant no. 929/17).
Taelin Karidi was partially supported by a fellowship from the Hebew University Center for Interdisciplinary Data Science Research.

\bibliography{anthology,custom}

\begin{thebibliography}{39}
\expandafter\ifx\csname natexlab\endcsname\relax\def\natexlab#1{#1}\fi

\bibitem[{Abend and Rappoport(2013)}]{abend-rappoport-2013-universal}
Omri Abend and Ari Rappoport. 2013.
\newblock \href {https://www.aclweb.org/anthology/P13-1023} {{U}niversal
  {C}onceptual {C}ognitive {A}nnotation ({UCCA})}.
\newblock In \emph{Proceedings of the 51st Annual Meeting of the Association
  for Computational Linguistics (Volume 1: Long Papers)}, pages 228--238,
  Sofia, Bulgaria. Association for Computational Linguistics.

\bibitem[{Ahmad et~al.(2019)Ahmad, Zhang, Ma, Hovy, Chang, and
  Peng}]{Ahmad2019OnDO}
Wasi~Uddin Ahmad, Zhisong Zhang, Xuezhe Ma, E.~Hovy, Kai-Wei Chang, and Nanyun
  Peng. 2019.
\newblock On difficulties of cross-lingual transfer with order differences: A
  case study on dependency parsing.
\newblock In \emph{NAACL-HLT}.

\bibitem[{Ammar et~al.(2016)Ammar, Mulcaire, Ballesteros, Dyer, and
  Smith}]{ammar-etal-2016-many}
Waleed Ammar, George Mulcaire, Miguel Ballesteros, Chris Dyer, and Noah~A.
  Smith. 2016.
\newblock \href {https://doi.org/10.1162/tacl_a_00109} {Many languages, one
  parser}.
\newblock \emph{Transactions of the Association for Computational Linguistics},
  4:431--444.

\bibitem[{Bugliarello and Okazaki(2020)}]{bugliarello-okazaki-2020-enhancing}
Emanuele Bugliarello and Naoaki Okazaki. 2020.
\newblock \href {https://doi.org/10.18653/v1/2020.acl-main.147} {Enhancing
  machine translation with dependency-aware self-attention}.
\newblock In \emph{Proceedings of the 58th Annual Meeting of the Association
  for Computational Linguistics}, pages 1618--1627, Online. Association for
  Computational Linguistics.

\bibitem[{Chen et~al.(2017)Chen, Huang, Chiang, and
  Chen}]{chen-etal-2017-improved}
Huadong Chen, Shujian Huang, David Chiang, and Jiajun Chen. 2017.
\newblock \href {https://doi.org/10.18653/v1/P17-1177} {Improved neural machine
  translation with a syntax-aware encoder and decoder}.
\newblock In \emph{Proceedings of the 55th Annual Meeting of the Association
  for Computational Linguistics (Volume 1: Long Papers)}, pages 1936--1945,
  Vancouver, Canada. Association for Computational Linguistics.

\bibitem[{Chen et~al.(2018)Chen, Wang, Utiyama, Sumita, and
  Zhao}]{Chen2018SyntaxDirectedAF}
Kehai Chen, Rui Wang, M.~Utiyama, E.~Sumita, and T.~Zhao. 2018.
\newblock Syntax-directed attention for neural machine translation.
\newblock In \emph{AAAI}.

\bibitem[{Devlin et~al.(2018)Devlin, Chang, Lee, and
  Toutanova}]{devlin2018bert}
Jacob Devlin, Ming-Wei Chang, Kenton Lee, and Kristina Toutanova. 2018.
\newblock Bert: Pre-training of deep bidirectional transformers for language
  understanding.
\newblock \emph{arXiv preprint arXiv:1810.04805}.

\bibitem[{Dozat and Manning(2017)}]{dozat2017biaffine}
Timothy Dozat and Christopher~D. Manning. 2017.
\newblock \href {https://openreview.net/forum?id=Hk95PK9le} {Deep biaffine
  attention for neural dependency parsing}.
\newblock In \emph{5th International Conference on Learning Representations,
  {ICLR} 2017, Toulon, France, April 24-26, 2017, Conference Track
  Proceedings}. OpenReview.net.

\bibitem[{Garc{\'\i}a(2006)}]{garcia2006mapping}
Noelia~Ram{\'o}n Garc{\'\i}a. 2006.
\newblock Mapping meaning onto form: A corpus-based contrastive study of
  nominal modification in english and spanish.
\newblock \emph{Languages in Contrast}, 6(2):307--334.

\bibitem[{Gardner et~al.(2018)Gardner, Grus, Neumann, Tafjord, Dasigi, Liu,
  Peters, Schmitz, and Zettlemoyer}]{gardner-etal-2018-allennlp}
Matt Gardner, Joel Grus, Mark Neumann, Oyvind Tafjord, Pradeep Dasigi,
  Nelson~F. Liu, Matthew Peters, Michael Schmitz, and Luke Zettlemoyer. 2018.
\newblock \href {https://doi.org/10.18653/v1/W18-2501} {{A}llen{NLP}: A deep
  semantic natural language processing platform}.
\newblock In \emph{Proceedings of Workshop for {NLP} Open Source Software
  ({NLP}-{OSS})}, pages 1--6, Melbourne, Australia. Association for
  Computational Linguistics.

\bibitem[{Hershcovich et~al.(2018)Hershcovich, Abend, and
  Rappoport}]{hershcovich2018multitask}
Daniel Hershcovich, Omri Abend, and Ari Rappoport. 2018.
\newblock \href {http://aclweb.org/anthology/P18-1035} {Multitask parsing
  across semantic representations}.
\newblock In \emph{Proc. of ACL}, pages 373--385.

\bibitem[{Liu et~al.(2020{\natexlab{a}})Liu, Zhou, Xu, Zheng, Chang, and
  Huang}]{liu-etal-2020-cross-lingual-dependency}
Lu~Liu, Yi~Zhou, Jianhan Xu, Xiaoqing Zheng, Kai-Wei Chang, and Xuanjing Huang.
  2020{\natexlab{a}}.
\newblock \href {https://doi.org/10.18653/v1/2020.findings-emnlp.265}
  {Cross-lingual dependency parsing by {POS}-guided word reordering}.
\newblock In \emph{Findings of the Association for Computational Linguistics:
  EMNLP 2020}, pages 2938--2948, Online. Association for Computational
  Linguistics.

\bibitem[{Liu et~al.(2020{\natexlab{b}})Liu, Hershcovich, Kranzlein, and
  Schneider}]{Liu2020LexicalSR}
Nelson~F. Liu, Daniel Hershcovich, Michael Kranzlein, and Nathan Schneider.
  2020{\natexlab{b}}.
\newblock Lexical semantic recognition.
\newblock \emph{ArXiv}, abs/2004.15008.

\bibitem[{Maniez(2012)}]{maniez2012corpus}
Fran{\c{c}}ois Maniez. 2012.
\newblock A corpus-based study of adjectival vs nominal modification in medical
  english.
\newblock In Alex Boulton, Shirley Carter-Thomas, and Elizabeth Rowley-Jolivet,
  editors, \emph{Corpus-Informed Research and Learning in {ESP}: {I}ssues and
  Applications}, pages 83--102. John Benjamins.

\bibitem[{Marneffe et~al.(2006)Marneffe, MacCartney, and
  Manning}]{stanford-dependency}
Marie-Catherine Marneffe, Bill MacCartney, and Christopher Manning. 2006.
\newblock Generating typed dependency parses from phrase structure parses.
\newblock volume~6.

\bibitem[{Meng et~al.(2019)Meng, Peng, and Chang}]{meng-etal-2019-target}
Tao Meng, Nanyun Peng, and Kai-Wei Chang. 2019.
\newblock \href {https://doi.org/10.18653/v1/D19-1103} {Target language-aware
  constrained inference for cross-lingual dependency parsing}.
\newblock In \emph{Proceedings of the 2019 Conference on Empirical Methods in
  Natural Language Processing and the 9th International Joint Conference on
  Natural Language Processing (EMNLP-IJCNLP)}, pages 1117--1128, Hong Kong,
  China. Association for Computational Linguistics.

\bibitem[{Minh Van~Nguyen and Nguyen(2021)}]{nguyen2021trankit}
Amir Pouran Ben~Veyseh Minh Van~Nguyen, Viet~Lai and Thien~Huu Nguyen. 2021.
\newblock Trankit: A light-weight transformer-based toolkit for multilingual
  natural language processing.
\newblock In \emph{Proceedings of the 16th Conference of the European Chapter
  of the Association for Computational Linguistics: System Demonstrations}.

\bibitem[{Mulcaire et~al.(2019)Mulcaire, Kasai, and
  Smith}]{mulcaire-etal-2019-polyglot}
Phoebe Mulcaire, Jungo Kasai, and Noah~A. Smith. 2019.
\newblock \href {https://doi.org/10.18653/v1/N19-1392} {Polyglot contextual
  representations improve crosslingual transfer}.
\newblock In \emph{Proceedings of the 2019 Conference of the North {A}merican
  Chapter of the Association for Computational Linguistics: Human Language
  Technologies, Volume 1 (Long and Short Papers)}, pages 3912--3918,
  Minneapolis, Minnesota. Association for Computational Linguistics.

\bibitem[{Nikolaev et~al.(2020)Nikolaev, Arviv, Karidi, Kenneth, Mitnik,
  Saeboe, and Abend}]{nikolaev-etal-2020-fine}
Dmitry Nikolaev, Ofir Arviv, Taelin Karidi, Neta Kenneth, Veronika Mitnik,
  Lilja~Maria Saeboe, and Omri Abend. 2020.
\newblock \href {https://doi.org/10.18653/v1/2020.acl-main.109} {Fine-grained
  analysis of cross-linguistic syntactic divergences}.
\newblock In \emph{Proceedings of the 58th Annual Meeting of the Association
  for Computational Linguistics}, pages 1159--1176, Online. Association for
  Computational Linguistics.

\bibitem[{Nivre et~al.(2016)Nivre, de~Marneffe, Ginter, Goldberg, Haji{\v{c}},
  Manning, McDonald, Petrov, Pyysalo, Silveira, Tsarfaty, and
  Zeman}]{nivre-etal-2016-universal}
Joakim Nivre, Marie-Catherine de~Marneffe, Filip Ginter, Yoav Goldberg, Jan
  Haji{\v{c}}, Christopher~D. Manning, Ryan McDonald, Slav Petrov, Sampo
  Pyysalo, Natalia Silveira, Reut Tsarfaty, and Daniel Zeman. 2016.
\newblock \href {https://www.aclweb.org/anthology/L16-1262} {{U}niversal
  {D}ependencies v1: A multilingual treebank collection}.
\newblock In \emph{Proceedings of the Tenth International Conference on
  Language Resources and Evaluation ({LREC}'16)}, pages 1659--1666,
  Portoro{\v{z}}, Slovenia. European Language Resources Association (ELRA).

\bibitem[{Nivre et~al.(2018)Nivre, Marongiu, Ginter, Kanerva, Montemagni,
  Schuster, and Simi}]{nivre-etal-2018-enhancing}
Joakim Nivre, Paola Marongiu, Filip Ginter, Jenna Kanerva, Simonetta
  Montemagni, Sebastian Schuster, and Maria Simi. 2018.
\newblock \href {https://doi.org/10.18653/v1/W18-6012} {Enhancing {U}niversal
  {D}ependency treebanks: A case study}.
\newblock In \emph{Proceedings of the Second Workshop on Universal Dependencies
  ({UDW} 2018)}, pages 102--107, Brussels, Belgium. Association for
  Computational Linguistics.

\bibitem[{Noh et~al.(2018)Noh, Han, Oh, and Kim}]{noh2018enhancing}
Youngbin Noh, Jiyoon Han, Tae~Hwan Oh, and Hansaem Kim. 2018.
\newblock Enhancing universal dependencies for korean.
\newblock In \emph{Proceedings of the second Workshop on Universal Dependencies
  (UDW 2018)}, pages 108--116.

\bibitem[{Ponti et~al.(2018)Ponti, Reichart, Korhonen, and
  Vuli{\'c}}]{ponti-etal-2018-isomorphic}
Edoardo~Maria Ponti, Roi Reichart, Anna Korhonen, and Ivan Vuli{\'c}. 2018.
\newblock \href {https://doi.org/10.18653/v1/P18-1142} {Isomorphic transfer of
  syntactic structures in cross-lingual {NLP}}.
\newblock In \emph{Proceedings of the 56th Annual Meeting of the Association
  for Computational Linguistics (Volume 1: Long Papers)}, pages 1531--1542,
  Melbourne, Australia. Association for Computational Linguistics.

\bibitem[{Rafaeli et~al.(2021)Rafaeli, Abend, Choshen, and
  Nikolaev}]{rafaeli2021speech}
Ofek Rafaeli, Omri Abend, Leshem Choshen, and Dmitry Nikolaev. 2021.
\newblock \href {http://arxiv.org/abs/2106.00745} {Part of speech and universal
  dependency effects on english arabic machine translation}.

\bibitem[{Rasooli and Collins(2019)}]{rasooli-collins-2019-low}
Mohammad~Sadegh Rasooli and Michael Collins. 2019.
\newblock \href {https://doi.org/10.18653/v1/N19-1385} {Low-resource syntactic
  transfer with unsupervised source reordering}.
\newblock In \emph{Proceedings of the 2019 Conference of the North {A}merican
  Chapter of the Association for Computational Linguistics: Human Language
  Technologies, Volume 1 (Long and Short Papers)}, pages 3845--3856,
  Minneapolis, Minnesota. Association for Computational Linguistics.

\bibitem[{Reddy et~al.(2017)Reddy, Täckström, Petrov, Steedman, and
  Lapata}]{reddy2017universal}
Siva Reddy, Oscar Täckström, Slav Petrov, Mark Steedman, and Mirella Lapata.
  2017.
\newblock \href {http://arxiv.org/abs/1702.03196} {Universal semantic parsing}.

\bibitem[{Schuster and Manning(2016)}]{Schuster2016EnhancedEU}
Sebastian Schuster and Christopher~D. Manning. 2016.
\newblock Enhanced english universal dependencies: An improved representation
  for natural language understanding tasks.
\newblock In \emph{LREC}.

\bibitem[{Schuster et~al.(2019)Schuster, Ram, Barzilay, and
  Globerson}]{schuster-etal-2019-cross}
Tal Schuster, Ori Ram, Regina Barzilay, and Amir Globerson. 2019.
\newblock \href {https://doi.org/10.18653/v1/N19-1162} {Cross-lingual alignment
  of contextual word embeddings, with applications to zero-shot dependency
  parsing}.
\newblock In \emph{Proceedings of the 2019 Conference of the North {A}merican
  Chapter of the Association for Computational Linguistics: Human Language
  Technologies, Volume 1 (Long and Short Papers)}, pages 1599--1613,
  Minneapolis, Minnesota. Association for Computational Linguistics.

\bibitem[{Stanovsky et~al.(2014)Stanovsky, Ficler, Dagan, and
  Goldberg}]{stanovsky-etal-2014-intermediary}
Gabriel Stanovsky, Jessica Ficler, Ido Dagan, and Yoav Goldberg. 2014.
\newblock \href {https://doi.org/10.3115/v1/W14-2413} {Intermediary semantic
  representation through proposition structures}.
\newblock In \emph{Proceedings of the {ACL} 2014 Workshop on Semantic Parsing},
  pages 66--70, Baltimore, MD. Association for Computational Linguistics.

\bibitem[{Tiktinsky et~al.(2020)Tiktinsky, Goldberg, and
  Tsarfaty}]{tiktinsky-etal-2020-pybart}
Aryeh Tiktinsky, Yoav Goldberg, and Reut Tsarfaty. 2020.
\newblock \href {https://doi.org/10.18653/v1/2020.acl-demos.7} {py{BART}:
  Evidence-based syntactic transformations for {IE}}.
\newblock In \emph{Proceedings of the 58th Annual Meeting of the Association
  for Computational Linguistics: System Demonstrations}, pages 47--55, Online.
  Association for Computational Linguistics.

\bibitem[{Wang and Eisner(2018{\natexlab{a}})}]{wang-eisner-2018-surface}
Dingquan Wang and Jason Eisner. 2018{\natexlab{a}}.
\newblock \href {https://doi.org/10.1162/tacl_a_00248} {Surface statistics of
  an unknown language indicate how to parse it}.
\newblock \emph{Transactions of the Association for Computational Linguistics},
  6:667--685.

\bibitem[{Wang and Eisner(2018{\natexlab{b}})}]{wang-eisner-2018-synthetic}
Dingquan Wang and Jason Eisner. 2018{\natexlab{b}}.
\newblock \href {https://doi.org/10.18653/v1/D18-1163} {Synthetic data made to
  order: The case of parsing}.
\newblock In \emph{Proceedings of the 2018 Conference on Empirical Methods in
  Natural Language Processing}, pages 1325--1337, Brussels, Belgium.
  Association for Computational Linguistics.

\bibitem[{Wang et~al.(2019)Wang, Che, Guo, Liu, and Liu}]{wang-etal-2019-cross}
Yuxuan Wang, Wanxiang Che, Jiang Guo, Yijia Liu, and Ting Liu. 2019.
\newblock \href {https://doi.org/10.18653/v1/D19-1575} {Cross-lingual {BERT}
  transformation for zero-shot dependency parsing}.
\newblock In \emph{Proceedings of the 2019 Conference on Empirical Methods in
  Natural Language Processing and the 9th International Joint Conference on
  Natural Language Processing (EMNLP-IJCNLP)}, pages 5721--5727, Hong Kong,
  China. Association for Computational Linguistics.

\bibitem[{Wolf et~al.(2020)Wolf, Debut, Sanh, Chaumond, Delangue, Moi, Cistac,
  Rault, Louf, Funtowicz, Davison, Shleifer, von Platen, Ma, Jernite, Plu, Xu,
  Le~Scao, Gugger, Drame, Lhoest, and Rush}]{wolf-etal-2020-transformers}
Thomas Wolf, Lysandre Debut, Victor Sanh, Julien Chaumond, Clement Delangue,
  Anthony Moi, Pierric Cistac, Tim Rault, Remi Louf, Morgan Funtowicz, Joe
  Davison, Sam Shleifer, Patrick von Platen, Clara Ma, Yacine Jernite, Julien
  Plu, Canwen Xu, Teven Le~Scao, Sylvain Gugger, Mariama Drame, Quentin Lhoest,
  and Alexander Rush. 2020.
\newblock \href {https://doi.org/10.18653/v1/2020.emnlp-demos.6} {Transformers:
  State-of-the-art natural language processing}.
\newblock In \emph{Proceedings of the 2020 Conference on Empirical Methods in
  Natural Language Processing: System Demonstrations}, pages 38--45, Online.
  Association for Computational Linguistics.

\bibitem[{Xu and Koehn(2021)}]{xu-koehn-2021-zero}
Haoran Xu and Philipp Koehn. 2021.
\newblock \href {https://www.aclweb.org/anthology/2021.adaptnlp-1.21}
  {Zero-shot cross-lingual dependency parsing through contextual embedding
  transformation}.
\newblock In \emph{Proceedings of the Second Workshop on Domain Adaptation for
  NLP}, pages 204--213, Kyiv, Ukraine. Association for Computational
  Linguistics.

\bibitem[{Yu et~al.(2015)Yu, Ji, Li, and Lin}]{yu-etal-2015-read}
Dian Yu, Heng Ji, Sujian Li, and Chin-Yew Lin. 2015.
\newblock \href {https://doi.org/10.3115/v1/N15-1126} {Why read if you can
  scan? trigger scoping strategy for biographical fact extraction}.
\newblock In \emph{Proceedings of the 2015 Conference of the North {A}merican
  Chapter of the Association for Computational Linguistics: Human Language
  Technologies}, pages 1203--1208, Denver, Colorado. Association for
  Computational Linguistics.

\bibitem[{Zeman et~al.(2017{\natexlab{a}})Zeman, Popel, Straka, Haji{\v{c}},
  Nivre, Ginter, Luotolahti, Pyysalo, Petrov, Potthast, Tyers, Badmaeva,
  Gokirmak, Nedoluzhko, Cinkov{\'a}, Haji{\v{c}}~jr., Hlav{\'a}{\v{c}}ov{\'a},
  Kettnerov{\'a}, Ure{\v{s}}ov{\'a}, Kanerva, Ojala, Missil{\"a}, Manning,
  Schuster, Reddy, Taji, Habash, Leung, de~Marneffe, Sanguinetti, Simi,
  Kanayama, de~Paiva, Droganova, Mart{\'\i}nez~Alonso, {\c{C}}{\"o}ltekin,
  Sulubacak, Uszkoreit, Macketanz, Burchardt, Harris, Marheinecke, Rehm,
  Kayadelen, Attia, Elkahky, Yu, Pitler, Lertpradit, Mandl, Kirchner, Alcalde,
  Strnadov{\'a}, Banerjee, Manurung, Stella, Shimada, Kwak, Mendon{\c{c}}a,
  Lando, Nitisaroj, and Li}]{zeman-etal-2017-conll}
Daniel Zeman, Martin Popel, Milan Straka, Jan Haji{\v{c}}, Joakim Nivre, Filip
  Ginter, Juhani Luotolahti, Sampo Pyysalo, Slav Petrov, Martin Potthast,
  Francis Tyers, Elena Badmaeva, Memduh Gokirmak, Anna Nedoluzhko, Silvie
  Cinkov{\'a}, Jan Haji{\v{c}}~jr., Jaroslava Hlav{\'a}{\v{c}}ov{\'a},
  V{\'a}clava Kettnerov{\'a}, Zde{\v{n}}ka Ure{\v{s}}ov{\'a}, Jenna Kanerva,
  Stina Ojala, Anna Missil{\"a}, Christopher~D. Manning, Sebastian Schuster,
  Siva Reddy, Dima Taji, Nizar Habash, Herman Leung, Marie-Catherine
  de~Marneffe, Manuela Sanguinetti, Maria Simi, Hiroshi Kanayama, Valeria
  de~Paiva, Kira Droganova, H{\'e}ctor Mart{\'\i}nez~Alonso,
  {\c{C}}a{\u{g}}r{\i} {\c{C}}{\"o}ltekin, Umut Sulubacak, Hans Uszkoreit,
  Vivien Macketanz, Aljoscha Burchardt, Kim Harris, Katrin Marheinecke, Georg
  Rehm, Tolga Kayadelen, Mohammed Attia, Ali Elkahky, Zhuoran Yu, Emily Pitler,
  Saran Lertpradit, Michael Mandl, Jesse Kirchner, Hector~Fernandez Alcalde,
  Jana Strnadov{\'a}, Esha Banerjee, Ruli Manurung, Antonio Stella, Atsuko
  Shimada, Sookyoung Kwak, Gustavo Mendon{\c{c}}a, Tatiana Lando, Rattima
  Nitisaroj, and Josie Li. 2017{\natexlab{a}}.
\newblock \href {https://doi.org/10.18653/v1/K17-3001} {{C}o{NLL} 2017 shared
  task: Multilingual parsing from raw text to {U}niversal {D}ependencies}.
\newblock In \emph{Proceedings of the {C}o{NLL} 2017 Shared Task: Multilingual
  Parsing from Raw Text to Universal Dependencies}, pages 1--19, Vancouver,
  Canada. Association for Computational Linguistics.

\bibitem[{Zeman et~al.(2017{\natexlab{b}})Zeman, Popel, Straka, Hajic, Nivre,
  Ginter, Luotolahti, Pyysalo, Petrov, Potthast, Tyers, Badmaeva, Gokirmak,
  Nedoluzhko, Cinkova, Hajic~jr., Hlavacova, Kettnerov\'{a}, Uresova, Kanerva,
  Ojala, Missil\"{a}, Manning, Schuster, Reddy, Taji, Habash, Leung,
  de~Marneffe, Sanguinetti, Simi, Kanayama, dePaiva, Droganova,
  Mart\'{i}nez~Alonso, \c{C}\"{o}ltekin, Sulubacak, Uszkoreit, Macketanz,
  Burchardt, Harris, Marheinecke, Rehm, Kayadelen, Attia, Elkahky, Yu, Pitler,
  Lertpradit, Mandl, Kirchner, Alcalde, Strnadov\'{a}, Banerjee, Manurung,
  Stella, Shimada, Kwak, Mendonca, Lando, Nitisaroj, and Li}]{conll2017shared}
Daniel Zeman, Martin Popel, Milan Straka, Jan Hajic, Joakim Nivre, Filip
  Ginter, Juhani Luotolahti, Sampo Pyysalo, Slav Petrov, Martin Potthast,
  Francis Tyers, Elena Badmaeva, Memduh Gokirmak, Anna Nedoluzhko, Silvie
  Cinkova, Jan Hajic~jr., Jaroslava Hlavacova, V\'{a}clava Kettnerov\'{a},
  Zdenka Uresova, Jenna Kanerva, Stina Ojala, Anna Missil\"{a}, Christopher~D.
  Manning, Sebastian Schuster, Siva Reddy, Dima Taji, Nizar Habash, Herman
  Leung, Marie-Catherine de~Marneffe, Manuela Sanguinetti, Maria Simi, Hiroshi
  Kanayama, Valeria dePaiva, Kira Droganova, H\'{e}ctor Mart\'{i}nez~Alonso,
  \c{C}a\u{g}rı \c{C}\"{o}ltekin, Umut Sulubacak, Hans Uszkoreit, Vivien
  Macketanz, Aljoscha Burchardt, Kim Harris, Katrin Marheinecke, Georg Rehm,
  Tolga Kayadelen, Mohammed Attia, Ali Elkahky, Zhuoran Yu, Emily Pitler, Saran
  Lertpradit, Michael Mandl, Jesse Kirchner, Hector~Fernandez Alcalde, Jana
  Strnadov\'{a}, Esha Banerjee, Ruli Manurung, Antonio Stella, Atsuko Shimada,
  Sookyoung Kwak, Gustavo Mendonca, Tatiana Lando, Rattima Nitisaroj, and Josie
  Li. 2017{\natexlab{b}}.
\newblock \href {http://www.aclweb.org/anthology/K/K17/K17-3001.pdf} {Conll
  2017 shared task: Multilingual parsing from raw text to universal
  dependencies}.
\newblock In \emph{Proceedings of the CoNLL 2017 Shared Task: Multilingual
  Parsing from Raw Text to Universal Dependencies}, pages 1--19, Vancouver,
  Canada. Association for Computational Linguistics.

\bibitem[{Zhang et~al.(2017)Zhang, Zhong, Chen, Angeli, and
  Manning}]{zhang2017tacred}
Yuhao Zhang, Victor Zhong, Danqi Chen, Gabor Angeli, and Christopher~D.
  Manning. 2017.
\newblock \href {https://nlp.stanford.edu/pubs/zhang2017tacred.pdf}
  {Position-aware attention and supervised data improve slot filling}.
\newblock In \emph{Proceedings of the 2017 Conference on Empirical Methods in
  Natural Language Processing (EMNLP 2017)}, pages 35--45.

\end{thebibliography}
\bibliographystyle{acl_natbib}

\appendix

\section{Additional Data on Edge Stability}
\label{sec:appendix_edge_stability}

In this section, we give additional details on the results reported in the \textit{Edge Stability and ZS Parsability} section (\S\ref{sec:edge_stability}).

\subsection{Edge Category Percentages and Standard Deviations}
\label{sec:edge_category_additional_data}

Proportions of each edge type in the PUD dataset in percentages are presented in Table~\ref{tab:edge_divergence_percentage}. Standard deviations for UD zero-shot performance scores for different edge categories are presented in Table~\ref{tab:edge_divergence_std}.

\begin{table*}[]
\small
\centering
\begin{tabular}{|l|c|c|c|c|c|c|}
\hline
\multicolumn{1}{|c|}{Edge Type} & \textbf{Russian} & \textbf{French} & \textbf{Chinese} & \textbf{Japanese} & \textbf{Korean} & \multicolumn{1}{l|}{\textbf{Arabic}} \\ \hline
\textbf{Fully Aligned} & 24 & 23 & 13 & 7 & 11 & 18 \\
\textbf{Partially Aligned} & 7.6 & 5.3 & 6.8 & 3.7 & 7.8 & 7.8 \\
\textbf{Unaligned} & 20 & 13 & 29 & 45 & 13 & 27 \\
\textbf{Misaligned} & 5 & 4 & 7 & 4.6 & 6.5 & 6 \\
\textbf{Flipped} & 1.6 & 1.3 & 2.4 & 1.6 & 3 & 2.6 \\ \hline
\textbf{Function Word} & 42 & 53 & 42 & 38 & 59 & 39 \\ \hline
\end{tabular}
\caption{Edge type proportions in the PUD dataset in percent. Rows corresponds to edge types, columns to PUD dataset languages. Details on the edge types can be found in \S\ref{sec:edge_stability}. The values do not sum up to 100\% due to rounding.}
\label{tab:edge_divergence_percentage}
\end{table*}

\begin{table*}[h!]
\small
\centering
\begin{tabular}{|l|ll|ll|ll|ll|ll|ll|}
\hline
\multicolumn{1}{|c|}{\textbf{}} & \multicolumn{2}{c|}{\textbf{Russian}} & \multicolumn{2}{c|}{\textbf{French}} & \multicolumn{2}{c|}{\textbf{Chinese}} & \multicolumn{2}{c|}{\textbf{Japanese}} & \multicolumn{2}{c|}{\textbf{Korean}} & \multicolumn{2}{c|}{\textbf{Arabic}} \\ \cline{2-13} 
\multicolumn{1}{|c|}{Edge Type} & \multicolumn{1}{c|}{UAS} & \multicolumn{1}{c|}{LAS} & \multicolumn{1}{c|}{UAS} & \multicolumn{1}{c|}{LAS} & \multicolumn{1}{c|}{UAS} & \multicolumn{1}{c|}{LAS} & \multicolumn{1}{c|}{UAS} & \multicolumn{1}{c|}{LAS} & \multicolumn{1}{c|}{UAS} & \multicolumn{1}{c|}{LAS} & \multicolumn{1}{l|}{UAS} & LAS \\ \hline
\textbf{Fully Aligned} & 0.75 & 0.94 & 1.02 & 1.24 & 1.34 & 2.15 & 2.62 & 2.24 & 1.36 & 1.13 & 1.66 & 1.77 \\
\textbf{Partially Aligned} & 1.61 & 1.71 & 1.18 & 0.82 & 1.55 & 1.5 & 1.91 & 1.69 & 1.04 & 1.63 & 2.27 & 1.62 \\
\textbf{Unaligned} & 1.4 & 1.09 & 0.85 & 0.9 & 1.41 & 1.25 & 2.77 & 0.47 & 1.46 & 1.15 & 1.24 & 1.15 \\
\textbf{Misaligned} & 1.13 & 1.03 & 1.21 & 1.21 & 0.8 & 0.85 & 1.22 & 0.93 & 1.42 & 1.04 & 1.15 & 0.98 \\
\textbf{Flipped} & 1.33 & 1.71 & 1.74 & 1.53 & 2.04 & 1.96 & 1.46 & 0.89 & 1.49 & 0.67 & 1.36 & 1.59 \\ \hline
\textbf{Func Word} & 0.73 & 0.61 & 0.41 & 0.69 & 1.05 & 0.71 & 1.32 & 1.06 & 1.67 & 1.5 & 0.87 & 1.02 \\ \hline
\end{tabular}
\caption{Standard deviations of UD zero-shot performance score per aligned edge type (averaged over 10 models). Rows corresponds to edge types; columns, to evaluation languages and score types. Details on the edge types can be found in \S\ref{sec:edge_stability}.}
\label{tab:edge_divergence_std}
\end{table*}

\subsection{Edge-Type composition of Different Stability Categories}
\label{sec:edge_category_supervised_appendix}

In order to gain better a insight in the edges in each of the stability categories, we analyze the performance of a supervised parser on these categories. We train 10 supervised models for each language, using the same UD parser as in \S\ref{sec:model_data}. For Russian, French, Chinese, Korean, and Japanese, we use the GSD corpora. For Arabic, we use the PADT corpus. We use the standard train-dev-test split and v2.5 of the UD dataset.

Results are presented in Table \ref{tab:edge_divergence_results_sup}, and standard deviations in Table \ref{tab:edge_divergence_std_sup}. We find that, surprisingly, the parser's performance on edges from different categories is generally ordered in the same order as in the ZS setting. This finding may suggest that cross-lingual stability is correlated with the ability of the parser to generalize within a language as well, a direction which we defer to future work. 

We compare the performance difference between the supervised and zero-shot parsers by first normalizing the performance on each category by dividing it by the performances of Fully Aligned edges, thus putting it range from 0 to 1, and then subtracting the normalized score of the supervised parser from the score of zero-shot parser, for each category.

The results are presented in Table \ref{tab:edge_divergence_normalized_res_diff_sup}. We find that the performance difference between the categories is more pronounced in the ZS setting than in the supervised setting. The only noticeable exceptions are in the Partially Aligned and Unaligned edges in Korean and in the Function Word edges in Arabic. We note that the Korean supervised parser displays very poor results, which is most likely due to annotation mismatches between the GSD and PUD Korean corpora.\footnote{Cf.\ the analysis by \citet{noh2018enhancing}.} It is also notable that Function-Word edges in Arabic show deviant results in the zero-shot settings as well.

\begin{table*}[h!]
\small
\centering
\begin{tabular}{|l|cc|cc|cc|cc|cc|cc|}
\hline
\multicolumn{1}{|c|}{\textbf{}} & \multicolumn{2}{c|}{\textbf{Russian}} & \multicolumn{2}{c|}{\textbf{French}} & \multicolumn{2}{c|}{\textbf{Chinese}} & \multicolumn{2}{c|}{\textbf{Japanese}} & \multicolumn{2}{c|}{\textbf{Korean}} & \multicolumn{2}{c|}{\textbf{Arabic}} \\ \cline{2-13} 
\multicolumn{1}{|c|}{Edge Type} & \multicolumn{1}{c|}{UAS} & LAS & \multicolumn{1}{c|}{UAS} & LAS & \multicolumn{1}{c|}{UAS} & LAS & \multicolumn{1}{c|}{UAS} & LAS & \multicolumn{1}{c|}{UAS} & LAS & \multicolumn{1}{l|}{UAS} & \multicolumn{1}{l|}{LAS} \\ \hline
\textbf{Fully Aligned} & 92 & 89 & 92 & 89 & 85 & 67 & 96 & 95 & 54 & 46 & 82 & 76 \\
\textbf{Partially Aligned} & 92 & 68 & 91 & 73 & 83 & 36 & 97 & 93 & 40 & 19 & 85 & 68 \\
\textbf{Unaligned} & 87 & 74 & 88 & 81 & 77 & 51 & 93 & 91 & 37 & 26 & 75 & 61 \\
\textbf{Misaligned} & 74 & 64 & 74 & 68 & 61 & 41 & 85 & 83 & 36 & 27 & 56 & 45 \\
\textbf{Flipped} & 80 & 65 & 83 & 71 & 66 & 50 & 94 & 89 & 36 & 25 & 70 & 48 \\ \hline
\textbf{Func Word} & 87 & 83 & 91 & 85 & 68 & 57 & 96 & 95 & 50 & 38 & 72 & 68 \\ \hline
\end{tabular}
\caption{UD supervised performance on the PUD corpora per edge type (averaged over 10 models) in percent. Rows corresponds to edge types; columns, to evaluation languages and score types. Details on the edge types can be found in \S\ref{sec:edge_stability}.}
\label{tab:edge_divergence_results_sup}
\end{table*}

\begin{table*}[h!]
\small
\centering
\begin{tabular}{|l|ll|ll|ll|ll|ll|ll|}
\hline
\multicolumn{1}{|c|}{} & \multicolumn{2}{c|}{\textbf{Russian}} & \multicolumn{2}{c|}{\textbf{French}} & \multicolumn{2}{c|}{\textbf{Chinese}} & \multicolumn{2}{c|}{Japanese} & \multicolumn{2}{c|}{Korean} & \multicolumn{2}{c|}{Arabic} \\ \cline{2-13} 
\multicolumn{1}{|c|}{Edge Type} & \multicolumn{1}{c|}{UAS} & \multicolumn{1}{c|}{LAS} & \multicolumn{1}{c|}{UAS} & \multicolumn{1}{c|}{LAS} & \multicolumn{1}{c|}{UAS} & \multicolumn{1}{c|}{LAS} & \multicolumn{1}{c|}{UAS} & \multicolumn{1}{c|}{LAS} & \multicolumn{1}{c|}{UAS} & \multicolumn{1}{c|}{LAS} & \multicolumn{1}{l|}{UAS} & LAS \\ \hline
\textbf{Fully Aligned} & 0.19 & 0.22 & 0.18 & 0.21 & 0.37 & 0.59 & 0.26 & 0.18 & 0.72 & 0.71 & 0.25 & 0.28 \\
\textbf{Partially Aligned} & 0.27 & 0.53 & 0.24 & 0.46 & 0.51 & 0.45 & 0.27 & 0.34 & 0.57 & 0.47 & 0.58 & 0.51 \\
\textbf{Unaligned} & 0.33 & 0.33 & 0.19 & 0.29 & 0.24 & 0.22 & 0.18 & 0.21 & 1.54 & 0.87 & 0.42 & 0.31 \\
\textbf{Misaligned} & 0.76 & 0.87 & 0.75 & 0.6 & 0.85 & 0.69 & 0.51 & 0.63 & 0.91 & 0.78 & 0.51 & 0.67 \\
\textbf{Flipped} & 0.74 & 0.85 & 0.24 & 0.53 & 0.94 & 1.25 & 0.38 & 0.49 & 0.99 & 0.81 & 0.87 & 0.63 \\ \hline
\textbf{Func Word} & 0.32 & 0.33 & 0.14 & 0.27 & 0.21 & 0.19 & 0.12 & 0.16 & 0.46 & 0.43 & 0.36 & 0.35 \\ \hline
\end{tabular}
\caption{Standard deviation of UD supervised performance scores per aligned edges type (averaged over 10 models). Rows corresponds to edge types; columns, to evaluation languages and score type. Details on the edge types can be found in \S\ref{sec:edge_stability}.
}
\label{tab:edge_divergence_std_sup}
\end{table*}

\begin{table*}[]
\small
\centering
\begin{tabular}{|l|ll|ll|ll|ll|ll|ll|}
\hline
\multicolumn{1}{|c|}{} & \multicolumn{2}{c|}{\textbf{Russian}} & \multicolumn{2}{c|}{\textbf{French}} & \multicolumn{2}{c|}{\textbf{Chinese}} & \multicolumn{2}{c|}{\textbf{Japanese}} & \multicolumn{2}{c|}{\textbf{Korean}} & \multicolumn{2}{c|}{\textbf{Arabic}} \\ \cline{2-13} 
\multicolumn{1}{|c|}{Edge Type} & \multicolumn{1}{c|}{UAS} & \multicolumn{1}{c|}{LAS} & \multicolumn{1}{c|}{UAS} & \multicolumn{1}{c|}{LAS} & \multicolumn{1}{c|}{UAS} & \multicolumn{1}{c|}{LAS} & \multicolumn{1}{c|}{UAS} & \multicolumn{1}{c|}{LAS} & \multicolumn{1}{c|}{UAS} & \multicolumn{1}{c|}{LAS} & \multicolumn{1}{l|}{UAS} & LAS \\ \hline
Fully Aligned & 0 & 0 & 0 & 0 & 0 & 0 & 0 & 0 & 0 & 0 & 0 & 0 \\
Partially Aligned & 0.07 & 0.26 & 0.04 & 0.14 & 0.05 & -0.07 & -0.01 & 0.3 & -0.23 & -0.18 & 0.11 & 0.45 \\
Unaligned & 0.12 & 0.12 & 0.07 & 0.09 & 0.18 & 0.19 & 0.39 & 0.69 & -0.07 & -0.06 & 0.14 & 0.16 \\
Misaligned & 0.19 & 0.17 & 0.17 & 0.17 & 0.26 & 0.27 & 0.47 & 0.58 & 0.04 & 0.23 & 0.1 & 0.1 \\
Flipped & 0.33 & 0.37 & 0.29 & 0.26 & 0.34 & 0.47 & 0.61 & 0.66 & -0.01 & 0.23 & 0.24 & 0.17 \\ \hline
Func Word & 0.18 & 0.18 & 0.08 & 0.07 & 0.21 & 0.34 & 0.61 & 0.65 & 0.26 & 0.31 & -0.09 & -0.26 \\ \hline
\end{tabular}
\caption{Comparison of the performance between the supervised and zero-shot parsers on each edge category. We first normalize the performance of each category by dividing it by the performances of Fully Aligned edges and then subtract the normalized score of the supervised parser from the score of zero-shot parser, for each category.
}
\label{tab:edge_divergence_normalized_res_diff_sup}
\end{table*}

\section{Translated Resources\footnote{All resources are available at \url{https://github.com/OfirArviv/translated_tacred}.}}
\label{sec:tranlated_resources_appendix}
\paragraph{Translated RE Datasets} In order for the Russian and Korean test sets to be representative of the different relation types in TACRED, we sampled the TACRED training set so that approximately 25\% of the examples are labeled as \textit{no\_relation} (in TACRED, 79.5\% are labeled as such) and the other 75\% are proportionally distributed between various relation types. We translated the examples using the Yandex\footnote{\url{https://translate.yandex.com/}} and Papago\footnote{\url{https://papago.naver.com/}} Translate APIs for Russian and Korean, respectively. Annotators, one for each target language, proficient in the target language and in English, then went over the translated examples, filtering out ones with low-quality translations and sampling others in their stead. They then manually annotated the entity spans of the relation participants corresponding to those identified annotated in English source sentences. The resulting Russian and Korean RE datasets consist of 533 parallel examples in both languages (and 3 additional examples in Russian), thus providing us with parallel, automatically translated, and manually curated and annotated RE datasets.

As the TACRED dataset is annotated with Stanford Dependencies \citep{stanford-dependency}, which are not designed to be cross-lingual, and not with UD, we use the Trankit \citep{nguyen2021trankit} supervised parser for parsing the datasets using the default provided pre-trained models. The resulting parses were checked and manually corrected by the annotators.

\paragraph{Translating Trigger words} The RE procedure we employ consults a list of trigger words collected for the different relations \citep{yu-etal-2015-read}.\footnote{The English trigger-word list is the same as in \citep{tiktinsky-etal-2020-pybart}.} We translate these trigger words to Korean and Russian in two ways. First, we automatically translate the entire word list using an automated machine translating system (Google Translate). This is not always sufficient, as in many cases there are multiple ways to translate a given trigger word. To remedy this, when annotating the translated TACRED subsample, we also record the spans of the trigger words in the translated sentences that correspond to those in the original English sentences.
\section{Model Hyperparameters}
\label{sec:appendix_models}
The hyperparameters of the UD parser are given in Table \ref{tab:zero_shot_hyperparams}.

\begin{table}[h]
\small
\centering
\begin{tabular}{|l|l|}
\hline
\multirow{2}{*}{\textbf{Input}} & Input dropout rate: 0.3 \\
 & Token embedder: \\ & bert-base-multilingual-cased \\ \hline
\multirow{7}{*}{\textbf{Encoder}} & Type: Stacked self attention \\
 & Input dim: 768 \\
 & Hidden dim: 400 \\
 & Projection dim: 512 \\
 & feedforward hidden dim: 400 \\
 & Layers \#: 3 \\
 & Attention heads \#: 8 \\ \hline
\multirow{5}{*}{\textbf{MLP and Attention}} & Arc MLP size: 500 \\
 & Label MLP size: 100 \\
 & MLP layers \#: 1 \\
 & Activation: ReLU \\
 & Dropout: 0.3 \\ \hline
\multirow{6}{*}{\textbf{Training}} & Batch size: 128 \\
 & Epochs \#: 100 \\
 & Early stopping: 50 \\
 & Adam lrate: 0.001 \\
 & Adam $\beta1$: 0.9 \\
 & Adam $\beta2$: 0.9 \\ \hline
\end{tabular}
\caption{Hyper-parameters used in the deep biaffine attention parser used in our experiments.
\label{tab:zero_shot_hyperparams}}
\end{table}

\section{Transformations}
\subsection{Normalization of Nominal Predicates}\label{sec:appendix_transformations}

One of the prominent sources of cross-lingual discrepancies in translation and when using UD is the handling of nominal predicates, i.e.\ nouns that evoke a semantic predicate-argument structure, sometimes also called a \textit{scene}. UD does not distinguish between scene-evoking nominal predicates and other nouns; see examples in \S\ref{ssec:nominal-predicate-normalization}.

In order to arrive at more cross-lingually stable handling of this construction, we employ UCCA \citep{abend-rappoport-2013-universal}, a semantic representation that explicitly distinguishes between scene-evoking and non-scene-evoking nouns. UCCA has two categories for Scene-evoking elements: States and Processes, of which we only consider the latter due to the mediocre parser performance on the former.

We use TUPA \citep{hershcovich2018multitask}, to parse the training corpus, identify subtrees headed by Processes and convert them to a clause-like form. Specifically, our transformation is as follows:
 \begin{itemize}
  \item \texttt{nsubj} (nominal subject) is converted to \texttt{csubj} (clausal subject).
  \item \texttt{nmod} (nominal modifier) is converted to \texttt{acl} (adnominal clause).
  \item \texttt{compound} is similar to \texttt{nmod} and thus is converted to \texttt{acl} as well.
  \item \texttt{obj} and \texttt{iobj} are converted to \texttt{ccomp} (complement clause).
  \item \texttt{obl} signify both indirect objects (\textit{Take money from a stranger}) and adverbial relations (\textit{Come back before tomorrow}) and therefore correspond to either \texttt{advcl} (adverbial clause) or \texttt{ccomp}. We convert them to \texttt{advcl} as adverbial-clause semantics in {\tt obl} is more common, according to \citet{nikolaev-etal-2020-fine}.
\end{itemize}

As we change the labels of nodes from nominal types to clausal types, we need to also change the labels of their dependents because, per UD convention, clause heads cannot be modified by adnominal dependents. We convert those nodes to their closest clause-level equivalents.
We adopted the following approach:

\begin{itemize}
  \item Adjectives and adverbs are often very close semantically (\textit{his play is magnificent} \ra\,\textit{he plays magnificently}), hence we convert \texttt{amod} to \texttt{advmod}.
  \item \texttt{acl} is converted to \texttt{advcl}.
  \item \texttt{nmod} and \texttt{compound} represent participants of a scene denoted by a nominal predicate. In a clause, a participant can be \texttt{nsubj}, \texttt{obj}, \texttt{iobj} or \texttt{obl} (cf.\ \textit{killing \textbf{of an artist}}: was the artist killed [obj] or did they kill somebody [nsubj]?). Choosing the correct participant type is a hard semantic task that we do not have good instruments to solve. Instead we introduce a new label, \texttt{A}, an undifferentiated participant, and convert \texttt{nmod} and \texttt{compound} to it.
\end{itemize}

The normalization of \texttt{nmod} and \texttt{compounds} into \texttt{A} creates a discrepancy: clauses transformed by our transformation contain \texttt{A}, while those were not still contain \texttt{nsubj}, \texttt{obj}, \texttt{iobj}, and \texttt{obl}. To harmonize this discrepancy we convert \texttt{nsubj}, \texttt{obj}, \texttt{iobj}, \texttt{obl} in all clauses into \texttt{A} as well. This results in a considerably simplified version of UD.

\section{Extrinsic Evaluation $p$-values}
\label{sec:appendix_p_val}

We use the paired bootstrap test to compute the $p$-values for the the differences between baseline scores and different transformation-based scores. For the {\sc Ensemble} setting, we find that all positive differences are significant ($<0.05$) and the vast majority are highly significant ($<0.001$). For the non-{\sc Ensemble} setting, we find that most are significant ($<0.05$). $p$-values are presented in Tables~\ref{tab:re_results_main_p_val} and~\ref{tab:re_results_denoise_p_val}.

 \begin{table*}[h!]
\small
\centering
\resizebox{.95\textwidth}{!}{
\begin{tabular}{|l|ccc|ccc|ccc|ccc|}
\hline
 & \multicolumn{6}{c|}{{\ul \textbf{Standard}}} & \multicolumn{6}{c|}{{\ul \textbf{Parallel}}} \\ \cline{2-13} 
 & \multicolumn{3}{c|}{\textbf{Korean}} & \multicolumn{3}{c|}{\textbf{Russian}} & \multicolumn{3}{c|}{\textbf{Korean}} & \multicolumn{3}{c|}{\textbf{Russian}} \\ \cline{2-13} 
\multicolumn{1}{|c|}{Trans.} & P & R & F1 & P & R & F1 & \multicolumn{1}{l}{P} & \multicolumn{1}{l}{R} & \multicolumn{1}{l|}{F1} & \multicolumn{1}{l}{P} & \multicolumn{1}{l}{R} & \multicolumn{1}{l|}{F1} \\ \hline
\sc{Nominal} & {0.000281} & {0.000235} & {0.000164} & 0.442 & {0.0153} & {0.0167} & {0.000303} & {0.00022} & {0.000213} & 0.415 & {0.000211} & {0.000181} \\
\sc{Predicate} & 0.193 & {0.0459} & {0.0483} &  0.878 & 0.954 & 0.953 & {0.0422} & {0.0808} & {0.0725} & 0.323 & {0.000108} & {0.000118} \\
\sc{Oblique} & {0.00164} & {0.0566} & {0.05} & 0.69 & 1 & 1 & {3.83e-05} & {0.0062} & {0.00225} & 0.345 & {1.54e-05} & {9.72e-06} \\\hline
\end{tabular}
}
\caption{$p$-values, computed using the paired bootstrap test, for the extrinsic evaluation of a transformed UD annotation on the pattern matching RE task in the {\sc Standard} and {\sc Parallel} settings. Columns and rows are the same as in Table \ref{tab:re_results_main}.}
\label{tab:re_results_main_p_val}
\end{table*}

 \begin{table*}[h!]
\small
\centering
\resizebox{.95\textwidth}{!}{
\begin{tabular}{|l|ccc|ccc|ccc|ccc|}
\hline
 & \multicolumn{6}{c|}{{\ul \textbf{Standard--Ensemble}}} & \multicolumn{6}{c|}{{\ul \textbf{Parallel--Ensemble}}} \\ \cline{2-13} 
 & \multicolumn{3}{c|}{\textbf{Korean}} & \multicolumn{3}{c|}{\textbf{Russian}} & \multicolumn{3}{c|}{\textbf{Korean}} & \multicolumn{3}{c|}{\textbf{Russian}} \\ \cline{2-13} 
\multicolumn{1}{|c|}{Trans.} & P & R & F1 & P & R & F1 & \multicolumn{1}{l}{P} & \multicolumn{1}{l}{R} & \multicolumn{1}{l|}{F1} & \multicolumn{1}{l}{P} & \multicolumn{1}{l}{R} & \multicolumn{1}{l|}{F1} \\ \hline
\sc{Nominal} & {7.47e-05} & {5.6e-05} & {4.01e-05} & {0.00262} & {5.34e-05} & {4.66e-05} & {0.000114} & {4.38e-05} & {6.21e-05} & {8.47e-05} & {6.13e-05} & {5.62e-05} \\
\sc{Predicate} & 0.522 & {4.14e-05} & {4.9e-05} &  1 & {4.49e-05} & {4.03e-05} & {5.55e-05} & {5.46e-05} & {6.41e-05} & {8.35e-05} &  {4.65e-05} & {3.43e-05} \\
\sc{Oblique} & {4.74e-05} & {4.88e-05} & {4.86e-05} & {5.35e-05} & {0.037} & {0.0188} & {4.02e-05} &  {4.09e-05} & {5.03e-05} & {0.000197} & {3.57e-05} & {3.29e-05} \\\hline
\end{tabular}
}
\caption{$p$-values, computed using the paired bootstrap test, extrinsic evaluation of a transformed UD annotation, using the pattern matching RE task, compared against the standard UD, for both Russian and Korean in the {\sc Ensemble} settings. Columns and rows are as in Table \ref{tab:re_results_main}.}
\label{tab:re_results_denoise_p_val}
\end{table*}

\end{document}